\documentclass[11pt]{article}
\usepackage{acl2015}
\usepackage{times}
\usepackage{latexsym}
\usepackage{amsmath}
\usepackage{amsfonts}
\usepackage{amssymb}
\usepackage{graphicx}
\usepackage{multicol}
\usepackage{hyperref}
\usepackage{cite}
\usepackage{url}

\newtheorem{theorem}{Theorem}

\title{A Fixed-Size Encoding Method for Variable-Length Sequences  \\ with its Application to  Neural Network Language Models}

\author{Shiliang Zhang$^1$, Hui Jiang$^2$, Mingbin Xu$^2$, Junfeng Hou$^1$, Lirong Dai$^1$ \\
  $^1$National Engineering Laboratory for Speech and Language Information Processing \\
  University of Science and Technology of China, Hefei, Anhui,  China\\
  $^2$Department of Electrical Engineering and Computer Science \\
  York University,  4700 Keele Street, Toronto, Ontario, M3J 1P3, Canada\\
  {\tt \small \{zsl2008,hjf176\}@mail.ustc.edu.cn, \{hj,xmb\}@cse.yorku.ca, lrdai@ustc.edu.cn}
}
\date{}

\begin{document}
\maketitle
\begin{abstract}
In this paper, we propose the new fixed-size ordinally-forgetting encoding (FOFE) method, which can almost uniquely encode any variable-length sequence of words into a fixed-size representation. FOFE can model the word order in a sequence using a simple ordinally-forgetting mechanism according to the positions of words. In this work, we have applied FOFE to feedforward neural network language models (FNN-LMs). Experimental results have shown that without using any recurrent feedbacks, FOFE based FNN-LMs can significantly outperform not only the standard fixed-input FNN-LMs but also the popular recurrent neural network (RNN) LMs.
\end{abstract}

\section{Introduction}

Language models play an important role in many applications like speech recognition, machine translation, information retrieval and nature language understanding. Traditionally, the back-off n-gram  models \cite{Katz.1987,Kneser.1995} are the standard approach to language modeling. Recently, neural networks have been successfully applied to language modeling and have achieved the state-of-the-art performance in many tasks. In neural network language models (NNLM), the feedforward neural networks (FNN) and recurrent neural networks (RNN) \cite{Elman1990} are two popular architectures. The basic idea of NNLMs is to use a projection layer to project discrete words into a continuous space and estimate word conditional probabilities in this space, which may be smoother to better generalize to unseen contexts. FNN language models (FNN-LM) \cite{Bengio2001,Bengio2003} usually use a limited history within a fixed-size context window to predict the next word. RNN language models (RNN-LM) \cite{Mikolov2010recurrent,Mikolov2012} adopt a time-delayed recursive architecture for the hidden layers to memorize the long-term dependency in language. Therefore, it is widely reported that RNN-LMs usually outperform FNN-LMs in language modeling.  While RNNs are theoretically powerful, the learning of RNNs needs to use the so-called back-propagation through time (BPTT) \cite{Werbos1990} due to the internal recurrent feedback cycles. The BPTT significantly increases the computational complexity of the learning algorithms and it may cause many problems in learning, such as gradient vanishing and exploding \cite{Bengio1994}. More recently, some new architectures have been proposed to solve these problems. For example, the long short term memory (LSTM) RNN \cite{Hochreiter1997} is an enhanced architecture to implement the recurrent feedbacks using various learnable gates, and it has obtained promising results on handwriting recognition \cite{Graves2009} and sequence modeling \cite{Graves2013}. 
Moreover, the so-called temporal-kernel recurrent neural networks (TKRNN) \cite{Sutskever2010}  have been proposed to handle the gradient vanishing problem. The main idea of TKRNN is to add direct connections between units in all time steps and every unit is implemented as an efficient leaky integrator, which makes it easier to learn the long-term dependency. Along this line, a temporal-kernel model has been successfully used for language modeling in \cite{Shi2013}.

Comparing with RNN-LMs, FNN-LMs can be learned in a simpler and more efficient way. However, FNN-LMs can not model the long-term dependency in language due to the fixed-size input window. In this paper, we propose a novel encoding method for discrete sequences, named {\em fixed-size ordinally-forgetting encoding} (FOFE), which can almost uniquely encode any variable-length word sequence into a fixed-size code. Relying on a constant forgetting factor, FOFE can model the word order in a sequence based on a simple ordinally-forgetting mechanism, which uses the position of each word in the sequence. Both the theoretical analysis and the experimental simulation have shown that FOFE can provide {\em almost} unique codes for variable-length word sequences as long as the forgetting factor is properly selected. In this work, we apply FOFE to neural network language models, where the fixed-size FOFE codes are fed to FNNs as input to predict next word, enabling FNN-LMs to model long-term dependency in language.  Experiments on two benchmark tasks, Penn Treebank Corpus (PTB) and Large Text Compression Benchmark (LTCB), have shown that FOFE-based FNN-LMs can not only significantly outperform the standard fixed-input FNN-LMs but also achieve better performance than the popular RNN-LMs with or without using LSTM. Moreover, our implementation also shows that FOFE based FNN-LMs can be learned very efficiently on GPUs without the complex BPTT procedure. 

\section{Our Approach: FOFE}
\label{sec.fofe}

Assume vocabulary size is $K$, NNLMs adopt the 1-of-K encoding vectors as input. In this case, each word in vocabulary is represented as a one-hot vector ${\mathbf e} \in \mathbb{R}^{K}$. The 1-of-K representation is a context independent encoding method. When the 1-of-K representation is used to model a word in a sequence, it can not model its history or context. 

\subsection{Fixed-size Ordinally Forgetting Encoding}

We propose a simple context-dependent encoding method for any sequence consisting of discrete symbols, namely {\em fixed-size ordinally-forgetting encoding} (FOFE). 
Given a sequence of words (or any discrete symbols), $S=\{ w_1, w_2, \cdots, w_T \}$, each word $w_t$ is first represented by a 1-of-K representation ${\bf e}_t$, from the first word $t=1$ to the end of the sequence $t=T$, FOFE encodes each partial sequence (history) based on a simple recursive formula (with $ {\bf z}_0 = {\bf 0}$) as:
\begin{equation}\label{eq-fofe-coding}
{\bf z}_t = \alpha \cdot {\bf z}_{t-1} + {\bf e}_{t}  \;\;\; (1 \leq t \leq T)
\end{equation}
where ${\bf z}_t$ denotes the FOFE code for the partial sequence up to $w_t$, and $\alpha$ ($0<\alpha<1$) is a constant forgetting factor to control the influence of the history on the current position. 
Let's take a simple example here, assume we have three symbols in vocabulary, e.g., {\em A}, {\em B}, {\em C}, whose 1-of-K codes are $[1,0,0]$, $[0,1,0]$ and $[0,0,1]$ respectively. In this case, the FOFE code for the sequence {\em \{ABC\}} is $[\alpha^2, \alpha ,1]$, and that of {\em \{ABCBC\}} is $[\alpha^4, \alpha+\alpha^3 ,1+\alpha^2]$.

Obviously, FOFE can encode any variable-length discrete sequence into a fixed-size code. Moreover, it is a recursive context dependent encoding method that smartly models the order information by various powers of the forgetting factor. Furthermore, FOFE has an appealing property in modeling natural languages that the far-away context will be gradually forgotten due to $\alpha<1$ and the nearby contexts play much larger role in the resultant FOFE codes.

\begin{figure}[t]
	\centering
	\includegraphics[width=0.9\linewidth]{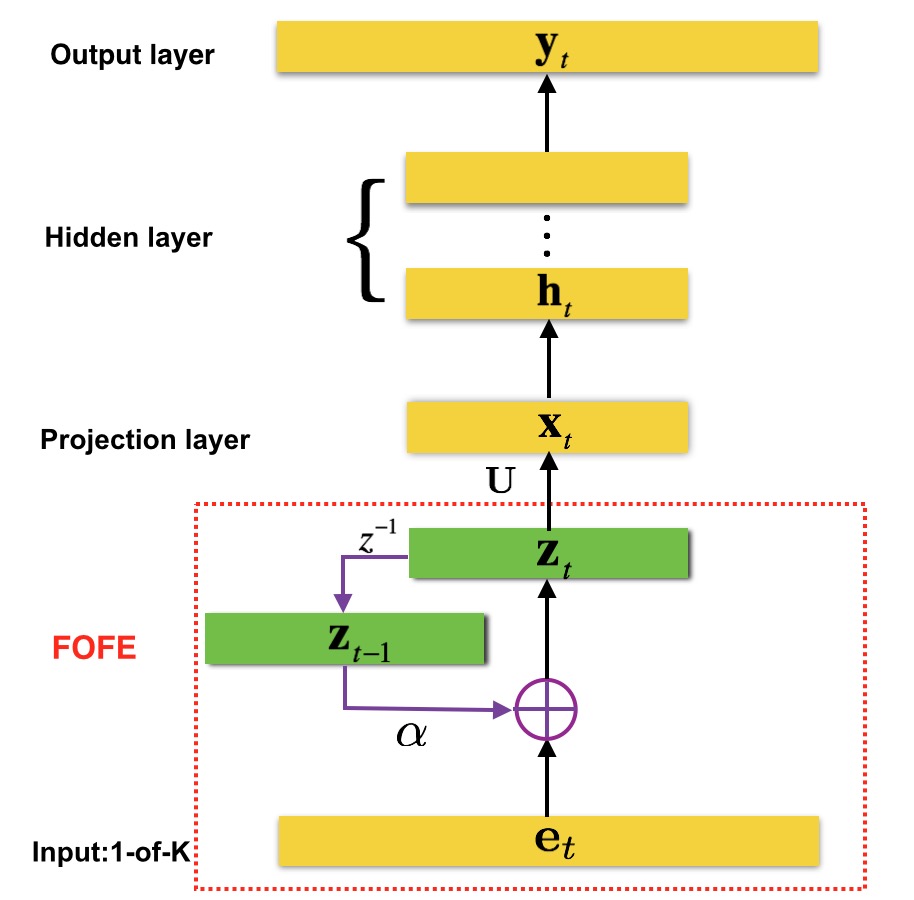}
	\caption{The FOFE-based FNN language model.}
	\label{fig:FOFE_bigram}
\end{figure}

\subsection{Uniqueness of FOFE codes}

Given the vocabulary (of $K$ symbols), for any sequence $S$ with a length of $T$, based on the FOFE code ${\bf z}_T$ computed as above, if we can always decode the original sequence $S$ unambiguously (perfectly recovering $S$ from ${\bf z}_T$), we say FOFE is unique. 

\begin{theorem}
\label{theorem-FOFE-alpha-less-half}
If the forgetting factor $\alpha$ satisfies $0<\alpha \leq 0.5$, {\em FOFE} is unique for any $K$ and $T$.	
\end{theorem}

The proof is simple because if the FOFE code has a value $\alpha^t$ in its $i$-th element, we may determine the word $w_i$ occurs in the position $t$ of $S$ without ambiguity since no matter how many times $w_i$ occurs in the far-away contexts ($<t$), they do not sum to $\alpha^t$ (due to $\alpha \leq 0.5$). If $w_i$ appears in any closer context ($>t$), the $i$-th element must be larger than $\alpha^t$.

For $0.5 < \alpha <1$, we have the following theorem:
\begin{theorem}
\label{theorem-FOFE-alpha-less-one}
For $0.5 < \alpha <1$, given any finite values of $K$ and $T$,  {\em FOFE} is almost unique everywhere for  $\alpha \in (0.5, 1.0)$, except only a finite set of countable choices of $\alpha$. 	
\end{theorem}

\begin{figure}[t]
	\centering
	\includegraphics[width=0.9\linewidth]{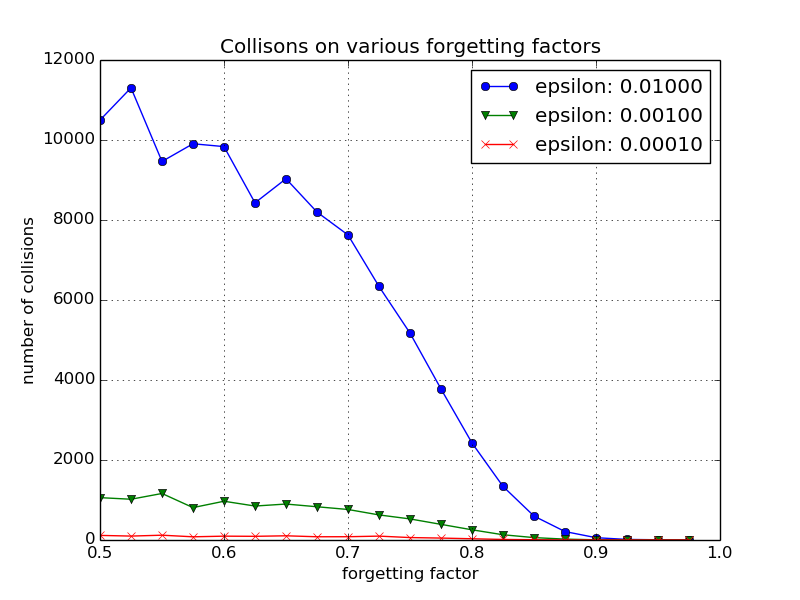}
	\caption{Numbers of collisions in simulation.}
	\label{fig:collisions}
\end{figure}

The complete proof \cite{Oguz2015} is given in Appendix A. Based on Theorem \ref{theorem-FOFE-alpha-less-one}, FOFE is unique almost everywhere between $(0.5, 1.0)$ only except a countable set of isolated choices of $\alpha$. In practice, the chance to exactly choose these isolated values between $(0.5, 1.0)$ is extremely slim, realistically almost impossible due to quantization errors in the system.   To verify this, 
we have run simulation experiments for all possible sequences up to $T=20$ symbols to count the number of collisions. Each collision is defined as the maximum element-wise difference between two FOFE codes (generated from two different sequences) is less than a small threshold $\epsilon$. 
In Figure \ref{fig:collisions}, we have shown the number of collisions (out of the total $2^{20}$ tested cases) for various $\alpha$ values when  $\epsilon=0.01$, $0.001$ and $0.0001$.\footnote{When we use a bigger value for $\alpha$, the magnitudes of the resultant FOFE codes become much larger. As a result, the number of collisions (as measured by a fixed absolute threshold $\epsilon$) becomes smaller.}
The simulation experiments have shown that the chance of collision is extremely small even when we allow a word to appear any times in the context. Obviously, in a natural language, a word normally does not appear repeatedly within a near context. Moreover, we have run the simulation to examine whether collisions actually occur in two real text corpora, namely PTB (1M words) and LTCB (160M words), using $\epsilon=0.01$, we have not observed a single collision for nine different $\alpha$ values between $[0.55, 1.0]$ (incremental $0.05$). 

\subsection{Implement FOFE for FNN-LMs}

\begin{figure}[t]
	\centering
	\includegraphics[width=0.9\linewidth]{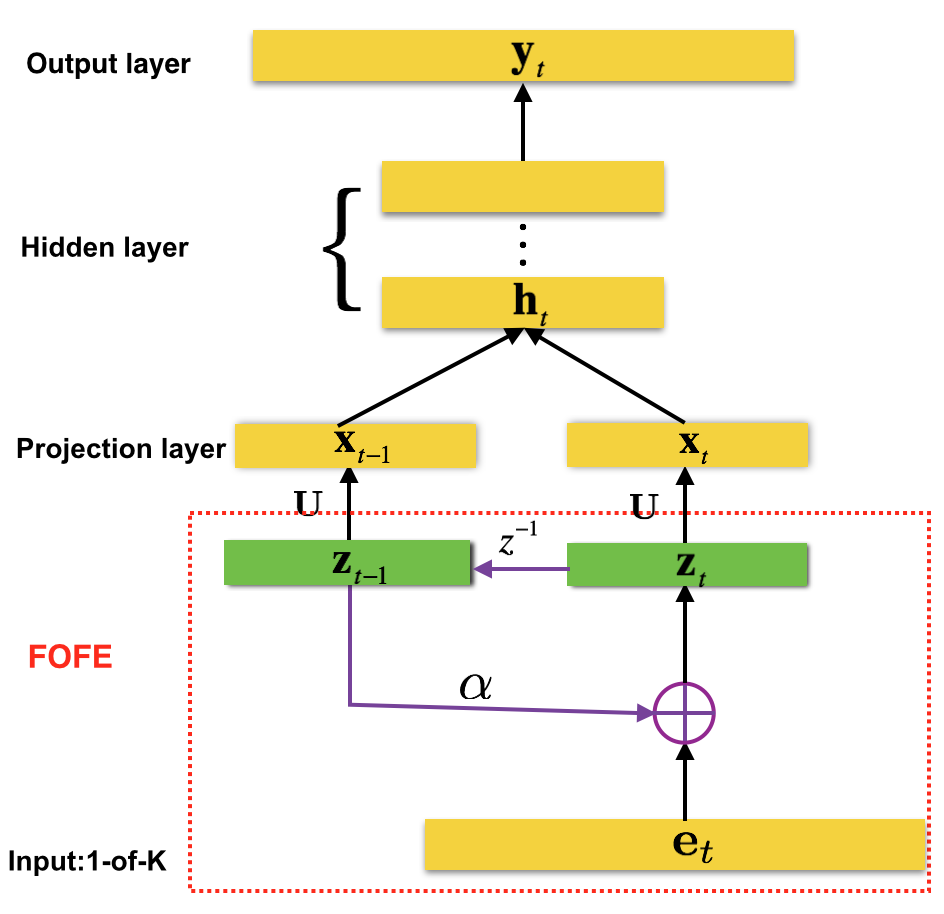}
	\caption{Illustration of a 2nd-order FOFE based FNN-LM.}
	\label{fig:FOFE_trigram}
\end{figure}

The architecture of a FOFE based neural network language model (FOFE-FNNLM) is as shown in Figure \ref{fig:FOFE_bigram}.  It is similar to standard bigram FNN-LMs except that it uses a FOFE code  
to feed into neural network LM at each time instance.
Moreover, the FOFE can be easily scaled to other n-gram based neural network LMs.
For example, 
Figure \ref{fig:FOFE_trigram} is an illustration of fixed-size ordinally forgetting encoding based tri-gram neural network language model.

FOFE is a simple recursive encoding method but a direct sequential implementation may not be efficient for the parallel computation platform like GPUs. 
Here, we will show that the FOFE computation can be efficiently implemented as sentence-by-sentence matrix multiplications, which are particularly suitable for the mini-batch based stochastic gradient descent (SGD) method running on GPUs. 

Given a sentence, $S = \{  {w}_1, {w}_2,   \cdots, {w}_T \} $, where each word is represented by a 1-of-K code as ${\bf e}_t$ $(1\leq t \leq T)$. The FOFE codes for all partial sequences in $S$ can be computed based on the following matrix multiplication:  
\begin{equation}\nonumber
{\bf S} = \left[ \begin{gathered}
1 \hfill \\
\alpha \quad 1 \hfill \\
{\alpha ^2}\;\;\alpha \quad 1 \hfill \\
\vdots \quad \;\;\quad  \ddots \quad 1 \hfill \\
{\alpha ^{T - 1}}\; \cdots \quad \alpha \quad 1 \hfill \\ 
\end{gathered}  \right] \left[ \begin{gathered}
{\bf e}_1 \hfill \\
{\bf e}_2 \hfill \\
{\bf e}_3 \hfill \\
\hspace{0.1cm} \vdots  \hfill \\
{\bf e}_T \hfill \\ 
\end{gathered}  \right] = {\bf M} {\bf V}
\end{equation}
where ${\bf V}$ is a matrix arranging all 1-of-K codes of the words in the sentence row by row, and ${\bf M}$ is a $T$-th order lower triangular matrix. Each row vector of ${\bf S}$ represents a FOFE code of the partial sequence up to each position in the sentence. 

This matrix formulation can be easily extended to a mini-batch consisting of several sentences.
Assume that a mini-batch is composed of N sequences, ${\cal L}=\{ S_1 \; S_2 \cdots S_N\}$, we can compute the FOFE codes for all sentences in the mini-batch as follows:
\begin{equation}\nonumber
{\bf \bar{S}} =\left[ \begin{gathered}
 {\bf M}_1 \hfill \\
\qquad  {\bf M}_2 \hfill \\
\qquad \qquad  \ddots  \hfill \\
\qquad \qquad \qquad  {\bf M}_N\hfill \\ 
\end{gathered}  \right]  \left[ \begin{gathered}
{\bf V}_1 \hfill \\
{\bf V}_2\hfill \\
\quad  \vdots  \hfill \\
{\bf V}_N \hfill \\ 
\end{gathered}  \right] = {\bf \bar{M}} {\bf \bar{V}}
\end{equation}

When feeding the FOFE codes to FNN as shown in Figure \ref{fig:FOFE_bigram}, we can compute the activation signals (assume $f$ is the activation function) in the first hidden layer for all histories in $S$  as follows: 
\begin{equation}
\label{eq.FOFE_forward}
{\bf H} = f\Big(({\bf M}  {\bf V}) {\bf U} {\bf W} + {\bf b}\Big) 
=  f\Big({\bf M} ({\bf V} {\bf U})  {\bf W} + {\bf b}\Big)  \nonumber
\end{equation}
where ${\bf U}$ denotes the word embedding matrix that projects the word indices onto a continuous low-dimensional continuous space. 
As above, ${\bf V} {\bf U}$ can be done efficiently by looking up the embedding matrix. Therefore, 
for the computational efficiency purpose, we may apply FOFE to the word embedding vectors  instead of the original high-dimensional one-hot vectors. In the backward pass, we can calculate the gradients with the standard back-propagation (BP) algorithm rather than BPTT. As a result, FOFE based FNN-LMs are the same as the standard FNN-LMs
in terms of computational complexity in training, which is much more efficient than RNN-LMs.

\section{Experiments}
\label{sec.Experiments}

We have evaluated the FOFE method for NNLMs on two benchmark tasks: i) the Penn Treebank (PTB) corpus of about 1M words, following the same setup as \cite{Mikolov2011Extension}.
The vocabulary size is limited to 10k. The preprocessing method and the way to split data into training/validation/test sets are the same as \cite{Mikolov2011Extension}. 
ii) The Large Text Compression Benchmark (LTCB) \cite{Mahoney2011}. In LTCB, we use the {\em enwik9} dataset, which is composed of  the first $10^9$ bytes of enwiki-20060303-pages-articles.xml. We  
split it into three parts:  training (153M), validation (8.9M) and testing (8.9M) sets. 
We limit the vocabulary size to 80k for LTCB and replace all out-of-vocabulary words by a $<$UNK$>$ token. 
Details of the two datasets can be found in Table \ref{tab:datasets}. \footnote{Matlab codes are available at \url{https://wiki.eecs.yorku.ca/lab/MLL/projects:fofe:start} for readers to reproduce all results reported in this paper.}   

\begin{table}[t]
	\centering
	\caption{The size of PTB and LTCB corpora in words. }
	\begin{tabular}{|c|c|c|c|}
		\hline  
		Corpus & train & valid & test \\ \hline  
		PTB & 930k & 74k &  82k \\ \hline
		LTC & 153M & 8.9M & 8.9M \\ 
		\hline 
	\end{tabular} 
	\label{tab:datasets}
\end{table}

\subsection{Experimental results on PTB }

We have first evaluated the performance of the traditional FNN-LMs, taking the previous several words as input, denoted as n-gram FNN-LMs here. We have trained neural networks with a linear projection layer (of 200 hidden nodes) and two hidden layers (of 400 nodes per layer). All hidden units in networks use the rectified linear activation function, i.e., $f(x)=\max(0,x)$. The nets are initialized based on the normalized initialization in \cite{Glorot2010}, without using any pre-training.  
We use SGD with a mini-batch size of 200 and an initial learning rate of 0.4. The learning rate is kept fixed as long as the perplexity on the validation set decreases by at least 1. After that, we continue six more epochs of training, where the learning rate is halved after each epoch. The performance (in perplexity) of various n-gram FNN-LMs is shown in Table \ref{tab:PTB_summary}. 

\begin{figure}[t]
	\centering
	\includegraphics[width=0.9\linewidth]{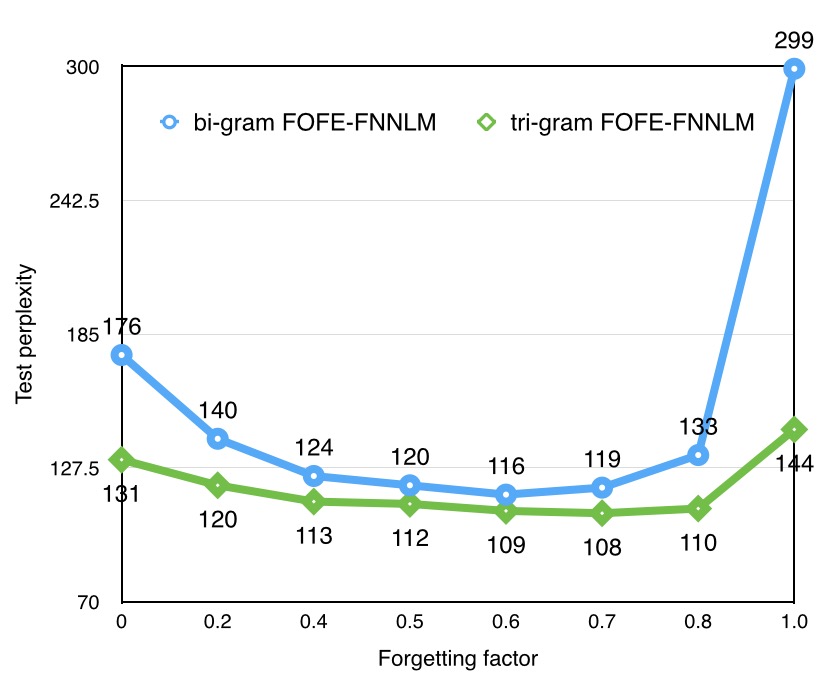}
	\caption{Perplexities of FOFE FNNLMs as a function of the forgetting factor.}
	\label{fig:Dif_factor}
\end{figure}

For the FOFE-FNNLMs, the net architecture and the parameter setting are the same as above. The mini-batch size is also 200 and each mini-batch is composed of several sentences up to 200 words (the last sentence may be truncated). All sentences in the corpus are randomly shuffled at the beginning of each epoch.  In this experiment, we first investigate how the forgetting factor $\alpha$ may affect the performance of LMs. We have trained two FOFE-FNNLMs: i) 1st-order (using ${\bf z}_t$ as input to FNN for each time $t$; ii) 2nd-order (using both ${\bf z}_t$ and ${\bf z}_{t-1}$ as input for each time $t$, with a forgetting factor varying between $[0.0,1.0]$. Experimental results in Figure \ref{fig:Dif_factor} have shown that
a good choice of $\alpha$ lies between $[0.5,0.8]$. 
Using a too large or too small forgetting factor will hurt the performance. A too small forgetting factor may limit the memory of the encoding while a too large $\alpha$ may confuse LM with a far-away history. 
In the following experiments, we set $\alpha=0.7$ for the rest experiments in this paper.

In Table \ref{tab:PTB_summary}, we have summarized the perplexities on the PTB test set for various models. The proposed FOFE-FNNLMs can significantly outperform the baseline FNN-LMs using the same architecture. 
For example, the perplexity of the baseline bigram FNNLM is 176, while the FOFE-FNNLM can improve to 116.  
Moreover, the FOFE-FNNLMs can even overtake a well-trained RNNLM (400 hidden units) in \cite{Mikolov2011Extension} and an LSTM in \cite{Graves2013}.  It indicates FOFE-FNNLMs can effectively model the long-term dependency in language without using any recurrent feedback. 
At last, the 2nd-order FOFE-FNNLM can provide further improvement, yielding the perplexity of 108 on PTB. It also outperforms all higher-order FNN-LMs (4-gram, 5-gram and 6-gram), which are bigger in model size. To our knowledge, this is one of the best reported results on PTB without model combination.
 
\begin{table}[t]
	\centering
	\caption{Perplexities on PTB for various LMs.}
	\begin{tabular}{|c|c|}
		\hline 
		Model  & Test PPL \\\hline \hline
		KN 5-gram \cite{Mikolov2011Extension}      &  141 \\
		FNNLM \cite{Mikolov2012}  & 140 \\
		RNNLM \cite{Mikolov2011Extension}  & 123 \\
		LSTM \cite{Graves2013}   &  117   \\ \hline
		bigram FNNLM & 176 \\
		trigram FNNLM & 131 \\
		4-gram FNNLM & 118 \\
		5-gram FNNLM & 114 \\
		6-gram FNNLM & 113 \\ \hline
		1st-order FOFE-FNNLM & 116 \\
		2nd-order FOFE-FNNLM & \textbf{108} \\ \hline
	\end{tabular} 
	\label{tab:PTB_summary}
\end{table}

\subsection{Experimental results on LTCB}

\begin{table}[t]
	\centering
	\caption{Perplexities on LTCB for various language models. [M*N] denotes the sizes of the input context window and projection layer.} 
	\begin{tabular}{|c|c|c|}
		\hline
		Model & Architecture & Test PPL \\\hline
		KN 3-gram &   -                &    156 \\
		KN 5-gram &  -                 &   132  \\\hline
		        & [1*200]-400-400-80k & 241 \\
		        & [2*200]-400-400-80k & 155\\
		FNN-LM  & [2*200]-600-600-80k & 150 \\
			    & [3*200]-400-400-80k & 131 \\
			    & [4*200]-400-400-80k & 125 \\ \hline
		RNN-LM   & [1*600]-600-80k       &   112 \\\hline
		              & [1*200]-400-400-80k & 120 \\
		FOFE    & [1*200]-600-600-80k &  115\\
		FNN-LM   & [2*200]-400-400-80k & 112\\
		              & [2*200]-600-600-80k  & {\bf 107} \\\hline

	\end{tabular} 
	\label{tab:WIKI_summary}
\end{table}

We have further examined the FOFE based FNN-LMs on a much larger text corpus, i.e. LTCB, which contains articles from Wikipedia. We have trained several  baseline systems: i) two n-gram LMs (3-gram and 5-gram) using the modified Kneser-Ney smoothing without count cutoffs; ii) several traditional FNN-LMs with different model sizes and input context windows (bigram, trigram, 4-gram and 5-gram ones); iii) an RNN-LM with one hidden layer of 600 nodes using the toolkit in \cite{Mikolov2010recurrent}, in which we have further used a spliced sentence bunch in \cite{Chen2014} to speed up the training on GPUs. Moreover, we have examined four FOFE based FNN-LMs with various model sizes and input window sizes (two 1st-order FOFE models and two 2nd-order ones). For all NNLMs, we have used an output layer of the full vocabulary (80k words). In these experiments, we have used an initial learning rate of 0.01, and a bigger mini-batch of 500 for FNN-LMMs and of 256 sentences for the RNN and FOFE models.
Experimental results in Table \ref{tab:WIKI_summary} have shown that the FOFE-based FNN-LMs can significantly outperform the baseline FNN-LMs (including some larger higher-order models) and also slightly overtake the popular RNN-based LM, yielding the best result (perplexity of 107) on the test set.

\section{Conclusions}
\label{sec.conclusion}

In this paper, we propose the fixed-size ordinally-forgetting encoding (FOFE) method to {\em almost} uniquely encode any variable-length sequence into a fixed-size code. In this work, FOFE has been successfully applied to neural network language modeling. 
Next, FOFE may be combined with neural networks \cite{Zhang2015a,Zhang2015b} for other NLP tasks, such as sentence modeling/matching, paraphrase detection, machine translation, question and answer and etc.

\section*{Acknowledgments}

This work was supported in part by the Science and Technology Development of Anhui Province, China (Grants No. 2014z02006) and the Fundamental Research Funds for the Central Universities from China, as well as an NSERC Discovery grant from Canadian federal govenment. We appreciate Dr. Barlas Oguz at Microsoft for his insightful comments and constructive suggestions on Theorem \ref{theorem-FOFE-alpha-less-one}.


\appendix 

\section*{Appendix A. The Proof of Theorem \ref{theorem-FOFE-alpha-less-one}}

{\bf Theorem 2:}
{\em For $0.5 < \alpha <1$, given any finite values of $K$ and $T$,  {\em FOFE} is almost unique everywhere for  $\alpha \in (0.5, 1.0)$, except only a finite set of countable choices of $\alpha$. }	

\textbf{Proof:}
When we decode a given FOFE code of an unknown sequence $S$ (assume the length of $S$ is not more than $T$), for any single value $1.0$ in the $i$-th position of the FOFE code, there are only two possible cases that may lead to ambiguity in decoding: (i) word $w_i$ appears in the current location of $S$; or (ii) word $w_i$ appears multiple times in the history of $S$ and the total contribution of them happens to be $1.0$. For case (ii) to happen, the forgetting factor $\alpha$ needs to satisfy at least one of the following polynomial equations: 
\begin{equation} \label{eq-FOFE-unique-formula}
	\sum_{t=1}^T \; \xi_t \cdot \alpha^t = 1.0
\end{equation}
where the above coefficients, $\xi_t$, are equal to either $1$ or $0$. If the word $w_i$ appears in the $t$-th location ahead in the history, we have $\xi_t=1$. Otherwise, $\xi_t=0$. We know, each equation 
in eq.(\ref{eq-FOFE-unique-formula}) is a $T$-th (or lower) order  polynomial equation. It can have at most $T$ real roots for $\alpha$. Moreover, since $\xi_t =\{ 0, 1\}$, we can only have a finite set of equations in eq.(\ref{eq-FOFE-unique-formula}). The total number is not more than $2^T$. Therefore, in total,  we can only have a finite number of $\alpha$ values that may satisfy at least one equation in eq.(\ref{eq-FOFE-unique-formula}), i.e., at most $T\cdot 2^T$ possible roots \cite{Oguz2015}. Among them, only a fraction of these roots lies between $(0.5, 1.0)$. Except these countable choices of $\alpha$ values, eq.(\ref{eq-FOFE-unique-formula}) never holds for any other $\alpha$ values between $(0.5, 1.0)$. As a result, case (ii) never happens in decoding except some isolated points of $\alpha$. This proves that the resultant FOFE code is {\em almost} unique between $(0.5, 1.0)$.    $\;\;\; \blacksquare$

\end{document}